\pdfoutput=1

\documentclass[11pt]{article}

\usepackage{ACL2023}

\usepackage{times}
\usepackage{latexsym}

\usepackage[T1]{fontenc}

\usepackage[utf8]{inputenc}

\usepackage{microtype}

\usepackage{inconsolata}

\usepackage{microtype}
\usepackage{float}
\usepackage{subcaption}
\usepackage{bm}
\usepackage{textcomp}
\usepackage{xurl}
\usepackage{booktabs}
\usepackage[inline]{enumitem}
\usepackage{amsfonts}
\usepackage{nicefrac}
\usepackage{microtype}
\usepackage{lipsum}
\usepackage{multirow}
\usepackage{makecell}
\usepackage{tabularx}
\usepackage{threeparttable}
\usepackage{colortbl}
\usepackage[separate-uncertainty=true]{siunitx}
\usepackage{datetime}
\usepackage{comment}
\usepackage{orcidlink}
\usepackage{cleveref}
\usepackage{microtype}
\usepackage[subtle]{savetrees}
\usepackage{algorithm}
\usepackage{algpseudocode}
\usepackage{placeins}
\graphicspath{ {./figures} }

\title{L1RA: Dynamic Rank Assignment in LoRA Fine-Tuning}

\author{Raul Singh\textsuperscript{*}{\normalfont ,} Nicolò Brunello\textsuperscript{*}{\normalfont ,}  Vincenzo Scotti\textsuperscript{\textdagger} \and Mark James Carman\textsuperscript{*}\\
  \textsuperscript{*}DEIB, Politecnico di Milano \\
  Via Ponzio 34/5, 20133, Milano (MI), Italy \\
  \textsuperscript{\textdagger}KASTEL, Karlsruhe Institute of Technology (KIT) \\
  Am Fasanengarten 5, 76131, Karlsruhe, Germany \\
  \texttt{raul.singh@mail.polimi.it} \qquad \texttt{nicolo.brunello@polimi.it} \\ 
  \texttt{vincenzo.scotti@kit.edu} \qquad \texttt{mark.carman@polimi.it}}

\begin{document}
\maketitle
\begin{abstract}
The ability of Large Language Models (LLMs) to solve complex tasks has made them crucial in the development of AI-based applications.
However, the high computational requirements to fine-tune these LLMs on downstream tasks pose significant challenges, particularly when resources are limited. 
In response to this challenge, we introduce \textsc{L1RA}, a novel technique aimed at dynamically distributing the rank of low-rank adapters during fine-tuning using \textsc{LoRA}. 
Given a rank budget (i.e., total sum of adapters rank), \textsc{L1RA} leverages $L_1$ regularisation to prune redundant ranks and redistribute them across adapters, thereby optimising resource utilisation. 
Through a series of comprehensive experiments, we empirically demonstrate that \textsc{L1RA} maintains comparable or even reduced computational overhead compared to other \textsc{LoRA} variants, including the vanilla approach, while achieving same or better performances. 
Moreover, the post-training analysis of rank distribution unveiled insights into the specific model components requiring the most adaptation to align with the task objective: the feed-forward layers and the attention output projection. 
These results highlight the efficacy of \textsc{L1RA} in not only enhancing the efficiency of LLM fine-tuning, but also in providing valuable diagnostic information for model refinement and customisation. 
In conclusion, \textsc{L1RA} stands as a promising technique for advancing the performance and interpretability of LLM adaptation, particularly in scenarios where computational resources are constrained.
\end{abstract}

\section{Introduction}

\emph{Large Language Models} (LLMs) have revolutionised \emph{Natural Language Processing} (NLP) and \emph{Artificial Intelligence} (AI)~\cite{DBLP:journals/corr/abs-2303-18223}, enabling sophisticated applications.
LLM's language understanding and generation capabilities make them suitable for an impressive number of applications~\cite{DBLP:journals/jmlr/RaffelSRLNMZLL20,DBLP:conf/nips/BrownMRSKDNSSAA20,DBLP:conf/iclr/SanhWRBSACSRDBX22}.
Moreover, their adoption as core for \emph{chatbots}~\cite{DBLP:journals/csur/ScottiST24} have made them essential for the final consumers of this technology.
However, to excel in these specific tasks, even conversation, LLMs often require \emph{fine-tuning}, a process essential for tailoring their vast pre-trained knowledge to new specific contexts and domains. 
This adaptation ensures optimal performance and task alignment, making fine-tuning a critical step in deploying LLMs effectively.

The fine-tuning process, however, presents challenges, particularly concerning computational resources. 
Adaptation to specific domains, such as chatbot dialogue or instruction-following tasks, demands substantial computational power, which may be impractical or unfeasible in resource-constrained environments. 
Recent advancements in efficient fine-tuning techniques, including \emph{Low-Rank Adaptation} (\textsc{LoRA})~\cite{DBLP:conf/iclr/HuSWALWWC22}, \emph{prefix tuning}~\cite{DBLP:conf/acl/LiL20} and the \emph{gradient-free methods} like \emph{Memory-efficient Zeroth-order Optimiser} (\textsc{MeZO})~\cite{DBLP:conf/nips/MalladiGNDL0A23}, offer promising solutions. 
These techniques leverage strategies like low-rank parameterisation to reduce computational overhead, making fine-tuning more accessible.

In this paper, we introduce \emph{$L_1$-regularised Rank Assignment} (\textsc{L1RA}): a technique aimed at enhancing the efficiency and effectiveness of LLM fine-tuning.
\textsc{L1RA} extends \textsc{LoRA} by introducing $L_1$ regularisation to enforce rank sparsity and dynamic rank allocation during training to get the best from the available resources.
Assuming a given \emph{rank budget} (i.e., total sum of \textsc{LoRA} adapter ranks), \textsc{L1RA} prunes redundant ranks and reallocates them across adapters during the fine-tuning process.
We pair \textsc{L1RA} with our tool \emph{Memory GPU Estimation of LLM Allocation for Training Optimisation} (\textsc{Memory-GELATO}) to be sure to match available resources constraints.
Through a series of experiments, ranging from small-scale analyses to comprehensive comparisons with other fine-tuning techniques, we evaluate the performance of \textsc{L1RA}.
The results highlight how \textsc{L1RA} can offer better comparable results to alternative \textsc{LoRA} variants reallocating ranks with negligible difference in resources consumption and better results even with respect to regular \textsc{LoRA}.

We divide the rest of the paper into the following sections.
In \Cref{sec:relatedworks}, we present the related works on efficient LLM fine-tuning.
In \Cref{sec:motivation}, we explain the reasons behind our work.
In \Cref{sec:l1ra,sec:appendixb}, we describe, respectively, the \textsc{L1RA} fine-tuning algorithm and the \textsc{Memory-GELATO} tool.
In \Cref{sec:evaluaton}, we outline the experiments to evaluate our model and in \Cref{sec:results} we present the obtained results.
In \Cref{sec:discussion} we comment on the results we obtained.
Finally, in \Cref{sec:conclusion}, we sum up our work and suggest possible future extensions.

\section{Related works}
\label{sec:relatedworks}

Efficient fine-tuning techniques have garnered increasing attention lately, due to the computational demands associated with adapting LLMs to specific tasks. 
The proposed techniques evolved significantly during the last few years.
Initial approaches like \emph{Transformer Adapters}~\cite {DBLP:conf/icml/HoulsbyGJMLGAG19,DBLP:conf/emnlp/BapnaF19} introduced additional parameters in the form of a pair of linear projections with a bottleneck in the middle, increasing network depth and latency, thereby hindering scalability. 
In response, \textsc{LoRA}-based solutions~\cite{DBLP:conf/iclr/HuSWALWWC22} have emerged as a promising alternative. 
\textsc{LoRA} addresses the limitations of adapters by introducing low-rank parameterisation, effectively reducing the number of parameters needed for adaptation. 
This technique has gained widespread adoption for its ability to achieve efficient fine-tuning without compromising performance. 
Alternative techniques like \textsc{MeZO}~\cite{DBLP:conf/nips/MalladiGNDL0A23} target the training algorithm rather than the network structure, focusing on fine-tuning through forward passes only, eliminating the need for backpropagation and the subsequent overhead.
Other approaches like prefix-tuning~\cite{DBLP:conf/acl/LiL20} learn only the embeddings of a \emph{continuos prompt} that can be used as a prefix to the input to condition the LLM output towards the desired task.
Among these techniques, \textsc{LoRA} stands out as the most adopted due to its effectiveness in balancing computational efficiency, performance and ease of use.

\begin{figure}[!ht] 
\begin{center}
\includegraphics[width=0.612\columnwidth]{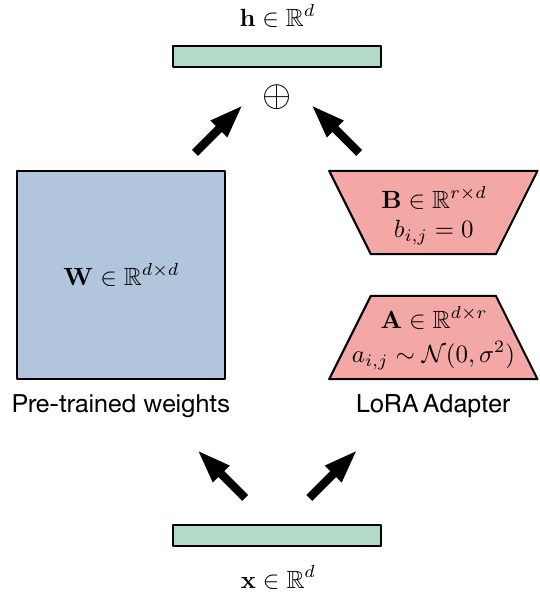}
\caption{\textsc{LoRA} adapters: pre-trained weights are frozen while the two adapter matrices are updated during the fine-tuning.}
\label{fig:lora}
\end{center}
\end{figure}

As premised, \textsc{LoRA} operates by introducing pairs of low-rank matrices $\mathbf{A} \in \mathbb{R}^{d_\textit{in} \times r}$ and $\mathbf{B} \in \mathbb{R}^{r \times d_\textit{out}}$ into the network architecture (see \Cref{fig:lora}); the product $\Delta\mathbf{W} \in \mathbb{R}^{d_\textit{in} \times d_\textit{out}}$ of these two matrices encodes the weights difference induced by fine-tuning for a specific weight matrix $\mathbf{W} \in \mathbb{R}^{d_\textit{in} \times d_\textit{out}}$ of the pre-trained model, as explained by \Cref{eq:lora}. 
During fine-tuning, these adapter matrices are updated while the original LLM parameters are kept frozen. 
By leveraging low-rank parameterisation, \textsc{LoRA} effectively reduces the computational overhead associated with fine-tuning while preserving the expressive power of the LLM. 
Moreover, this approach has demonstrated empirically significant improvements in efficiency without sacrificing performance across various downstream tasks.

\vspace{-1em}
\begin{equation}
\label{eq:lora}
    \mathbf{h} = \mathbf{x} \cdot \left(\mathbf{W} + \Delta\mathbf{W}\right) = \mathbf{x} \cdot \mathbf{W} + \mathbf{x} \cdot \left(\mathbf{A} \cdot \mathbf{B}\right)
\end{equation}

While \textsc{LoRA} offers notable benefits, several variants have been proposed to address specific limitations or further enhance its capabilities. 
Examples of these alternative solutions are those aimed at stabilising the training process, like \textsc{LoRA+}~\cite{DBLP:journals/corr/abs-2402-12354}, which introduces a matrix-specific scaling parameter on the learning rate to improve performances and convergence time, and \emph{Rank-Stabilised} \textsc{LoRA} (\textsc{rsLoRA})~\cite{DBLP:journals/corr/abs-2312-03732}, which uses a rank correcting factor to prevent gradient collapse.
Other variants, like \emph{Quantised} \textsc{LoRA} (\textsc{QLoRA})~\cite{DBLP:conf/nips/DettmersPHZ23}, aim at further reducing computational complexity by heavily quantising the base model weights (for reduced memory requirements and increased inference speed) while operating on floating-point representation of the trainable weights (for numeric precision and, thus, training stability), thereby improving the overall efficiency. 
In this paper we focus on techniques for adaptive rank allocation.
In fact, \textsc{LoRA} adapters depend on the rank hyper-parameter, which can be selected dynamically for each pair of adapter matrices.
In this sense, some solutions have been proposed to tackle the issue of rank selection in order to
\begin{enumerate*}[label=(\roman*)]
  \item get rid of unused parameters and
  \item find the best possible rank allocation allowed by the available memory.
\end{enumerate*}

One of the first solutions for dynamic rank allocation was presented with \textsc{AdaLoRA}~\cite{DBLP:conf/iclr/ZhangCBH0CZ23}, which enforces a \emph{Singular Value Decomposition}-inspired (SVD-inspired) decomposition of the adapter weights through additional regularisation terms in the loss. 
Further refinements of this technique came with \emph{Sparse} \textsc{LoRA} (\textsc{SoRA})~\cite{DBLP:conf/emnlp/DingLWCZL023}, which uses an intermediate gating mechanism with $L_1$ regularisation and \emph{proximal gradient descent} to iteratively reduce the used ranks, and, \emph{Vector-based Random matrix Adaptation} (\textsc{VeRA})~\cite{DBLP:journals/corr/abs-2310-11454}, which reduces the trainable \textsc{LoRA} parameters through shared random weights matrices and works on rank allocation updating only layer-specific parameter vectors.
In parallel, \emph{Dynamic rank selection} \textsc{LoRA} (\textsc{DyLoRA})~\cite{DBLP:conf/eacl/ValipourRKG23}, proposed a solution exploring a range of possible ranks during training to find the optimal ones for each matrix.

\section{Motivations}
\label{sec:motivation}

\textsc{LoRA} adapters represent a valuable step towards end-user fine-tuning of LLMs, making this technology more accessible and customisable.
The existence of techniques like \textsc{AdaLoRA}, \textsc{SoRA} or \textsc{DyLoRA} allowing for dynamic rank and pruning (i.e., removing the $i$-th column in $\mathbf{A}$ and the $i$-th row in $\mathbf{B}$) are the results of advances towards better exploitation of computational resources.
Hereafter, we highlight some points of improvement for \textsc{AdaLoRA} and \textsc{SoRA} (the main solutions for dynamic rank allocation), in terms of computational resources exploitation, that are motivating our work. 

\textsc{AdaLoRA} proposes a SVD-inspired formulation of the adapter:

\vspace{-.5em}
\begin{equation}
\Delta\mathbf{W} = \mathbf{U} \cdot \bm{\Sigma} \cdot \mathbf{V}^\top = \mathbf{U} \cdot \mathrm{diag}(\bm{\sigma}) \cdot \mathbf{V}^\top
\end{equation}

\noindent where $\mathbf{U} \in \mathbb{R}^{(d_\textit{in} \times r)}$, $\mathbf{V} \in \mathbb{R}^{(d_\textit{out} \times r)}$, $\bm{\sigma} \in {\mathbb{R}_0^+}^r$.
Then, it enforces an additional regularisation term $\mathcal{L}_\textit{SVD}(\Delta\mathbf{W})$ to the loss to imposing orthonormality on the adapter matrices.

\vspace{-.5em}
\begin{equation}
\Delta\mathbf{W} = \|\mathbf{U}^\top \cdot \mathbf{U} - \mathbf{I}\|_2^2 + \|\mathbf{V}^\top \cdot \mathbf{V} - \mathbf{I}\|_2^2
\end{equation}

Despite this constraint allows to interpret the values of $\bm{\sigma}$ as the eigenvalues and, thus, prune all elements corresponding to null eigenvalues in increases the memory and time requirements of the training process with respect to a normal \textsc{LoRA}.

\textsc{SoRA} builds on top of \textsc{AdaLoRA}, discarding the SVD constraint and substituting the vector of eigenvalues with a gating vector $\mathbf{g} \in \mathbb{R}^r$ and enforcing sparsity adding to the loss a L$1$ regularisation penalty on $g$.
This simple, yet effective solution, encourages to prune all elements corresponding to a $0$ valued element in the gate, as they will be ignored in the computation of the output (exactly as the elements corresponding to a null eigenvalue).
The complete formulation of \textsc{SoRA} includes the proximal
gradient update using a thresholding function that ensures training stability. 
This addition is already part of the optimiser we use in our experiments (see \Cref{sec:appendixc} for further details).

\begin{figure}[!ht] 
\begin{center}
\includegraphics[width=.924\columnwidth]{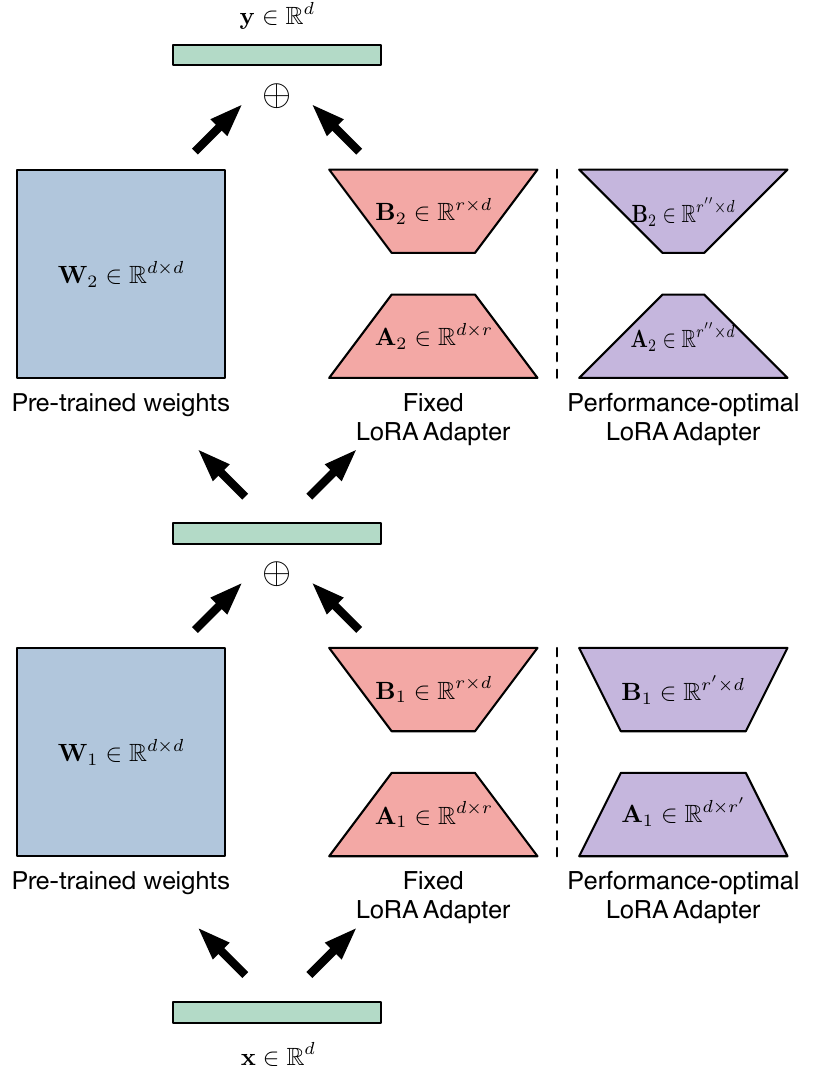}
\caption{Motivating example: $r'$ and $r''$ are such that $r' + r'' = 2 r$, so  that the total amount of adapters memory is the same with and without optimal allocation.}
\label{fig:motivatingex}
\end{center}
\end{figure}

All the proposed solutions for dynamic rank assignment correctly work to reduce the rank used in the adapter matrices.
However, these \textsc{LoRA} variants are limited in the sense that they do not allow for spare (unused) ranks re-assignment and they rather wait for the end of training to prune the matrices. 
They instead propose starting directly from higher ranks, usually $\nicefrac{3r}{2}$, which increase the overall memory requirement with respect to a base \textsc{LoRA} operating with the same resources and rank $r$.
Consider the toy example in \Cref{fig:motivatingex}, where we have the comparison between the matrices of \textsc{LoRA} with fixed rank allocation and the matrices with performance-optimal rank allocation.
In this case we would have a rank budget of $2 r$ that, in the performance-optimal allocation, is divided between $r'$, in the first adapter, and $r''$, in the second adapter, that $r' + r'' = 2 r$ and $r' > r > r''$.
In this configuration, with adapters like \textsc{AdaLoRA} or \textsc{SoRA}, we would need to start at least from a rank budget of $2 r' > 2 r$ to reach the performance optimal allocation, which is above the available budget of $2 r$.
Moreover, it may be the case where, since we are talking of constrained resources, the model with all the adapters starting from rank $r'$ would not fit in memory.

Besides the theoretical aspects of staying within the rank budget, we also have a ``physical'' constraint given by the amount of available GPU memory.
To tackle this problem we developed the \textsc{Memory-GELATO} tool, which comes as a complement to \textsc{L1RA}.
Though accurate estimates of the memory usage we can identify the starting rank without exceeding the available resources.
Similarly to other solutions, \textsc{L1RA} can drop the ranks in excess, but differently from the other takes care of re-allocating at runtime those ranks, all of this staying within the given budget.

In this section we described exactly the problems we tackle with our work: \emph{how to get the best performances given a fixed rank or memory budget}? 
In other words, our contribution is an algorithm that dynamically re-allocates rank amongst adapter matrices in order to maximise performance given a fixed maximum memory budget available, complemented with a tool for memory budget estimation.



\section{\textsc{L1RA}}
\label{sec:l1ra}

\begin{figure}[!ht] 
\begin{center}
\includegraphics[width=0.754\columnwidth]{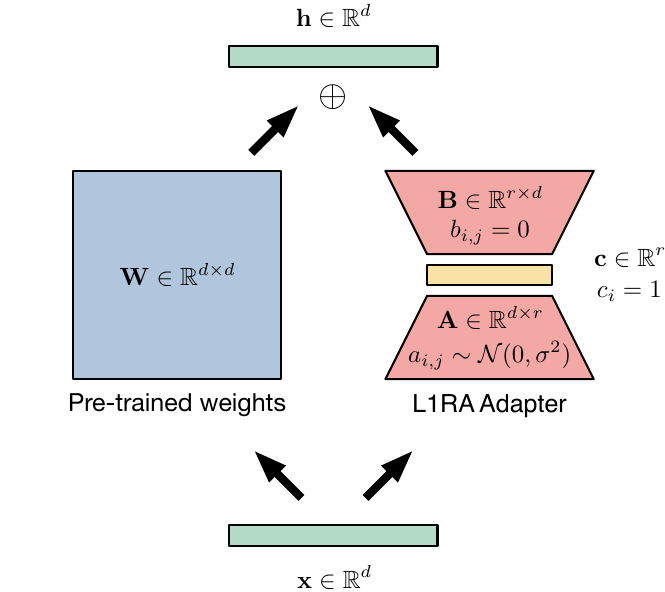}
\caption{\textsc{L1RA} adapters}
\label{fig:l1raadapter}
\end{center}
\vspace{-1.5em}
\end{figure}

\textsc{L1RA} adapters, depicted in \Cref{fig:l1raadapter}, extend the \textsc{LoRA} framework by introducing rank pruning and reallocation mechanisms within a fixed rank or memory budget. 
The goal of \textsc{L1RA} is to identify the performance-optimal rank configuration in computational constrained settings where memory --and time-- may be limited. 
This dynamic rank adjustment ensures that the model efficiently utilises the available resources, enhancing performance without exceeding the same constraints a vanilla \textsc{LoRA} adapter would have.

Mathematically, given an input vector $\mathbf{x} \in \mathbb{R}^{d_\textit{in}}$, we compute the output $\mathbf{h} \in \mathbb{R}^{d_\textit{out}}$ of a \textsc{L1RA} adapter as described in \Cref{eq:l1ra}. 
Where $\mathbf{W} \in \mathbb{R}^{d_\textit{in} \times d_\textit{out}}$ is the original matrix of pre-trained weights, $\Delta\mathbf{W} \in \mathbb{R}^{d_\textit{in} \times d_\textit{out}}$ is the adapter matrix of weights decomposed in $\mathbf{A} \in \mathbb{R}^{d_\textit{in} \times r}$, $\mathbf{c} \in \mathbb{R}^r$ and $\mathbf{B} \in \mathbb{R}^{r \times d_\textit{out}}$, and the $\mathrm{diag}(\cdot)$ function outputs a diagonal matrix with the elements of the input vector as values on the diagonal.
The rank $r$ depends on the specific adapter and is selected through optimisation during training.

\vspace{-1em}
\begin{equation}
\label{eq:l1ra}
    \mathbf{h} = \mathbf{x} \cdot \left(\mathbf{W} + \Delta\mathbf{W}\right) = \mathbf{x} \cdot \mathbf{W} + \mathbf{x} \cdot \left(\mathbf{A} \cdot \mathrm{diag}(\mathbf{c}) \cdot \mathbf{B}\right)
\end{equation}

The $\mathbf{c}$ vector we introduced, similar to the gating system of \textsc{SoRA}, is a technical device to ease the enforcing of the sparsity constraint.
We could have obtained the same effect of imposing sparsity on $\mathbf{c}$ (resulting from the regularisation term $\lambda\|\mathbf{c}\|_1$), by applying the same $L_1$ constraint to the columns of the input projection matrix $\mathbf{A}$ of a regular \textsc{LoRA} adapter. 
Doing so would have resulted, however, in much slower redistribution of rank, since an entire column of $\mathbf{A}$ would need to have converged to zero before it could be removed and reassigned to a different matrix, whereas a single component of the $\mathbf{c}$ vector falling to zero is sufficient for reassignment. 
Thus, while the $\mathbf{c}$ vector does introduce a small number of additional parameters, it results in faster and more direct rank-sparsification, while not affecting the overall transformation of the adapter. 

\begin{algorithm}
\caption{\textsc{L1RA} pseudocode}\label{alg:trainalg}

\begin{algorithmic}

\footnotesize

\Require \\
\begin{itemize}
    \item $\vartheta$ \Comment{Model parameters}
    \item $\mathcal{D}$ \Comment{Data}
    \item $r \in \mathbf{N^+}$ \Comment{Initial adapters rank}
\end{itemize}

\State $\Delta\vartheta \gets \{\}$ \Comment{Adapter parameters}
\For{$\mathbf{W} \in \vartheta$} \Comment{Initialise adapters of all layers}
    \State $\mathbf{A} \gets \mathbf{A} \in \mathbb{R}^{d \times r} \sim \mathcal{N}(0, \sigma^2)$
    \State $\mathbf{B} \gets \mathbf{0} \in \{0\}^{r \times d}$
    \State $\mathbf{c} \gets \mathbf{1} \in \{1\}^r$
    \State $\Delta\vartheta \gets \Delta\vartheta \cup \{(\mathbf{A}, \mathbf{B}, \mathbf{c})\}$
\EndFor

\For{$i \in [0, n_\textit{epochs}) \subseteq \mathbb{N}$}  \Comment{Iterate over epochs}
    \For{$X \in \mathcal{D}$} \Comment{Iterate over training samples}
        \State $\mathcal{L}(\Delta\vartheta) \gets -\ln{P(X; \vartheta, \Delta\vartheta)} + \lambda \cdot \sum_{(\mathbf{A}, \mathbf{B}, \mathbf{c}) \in \Delta\vartheta}{\|\mathbf{c}\|_1}$ \Comment{Get loss}
        \State $\Delta\vartheta \gets \Delta\vartheta - \eta  \cdot \nabla_{\Delta\vartheta} \cdot \mathcal{L}(\Delta\vartheta)$ \Comment{Update adapter weights}
        \State $\rho \gets 0$ \Comment{Initialise spare ranks}
        \State $\Delta\vartheta_u \gets []$ \Comment{Initialise list of unpruned adapters}
        \For{$(\mathbf{A}, \mathbf{B}, \mathbf{c}) \in \Delta\vartheta$} \Comment{Iterate over adapters}
            \If{$\exists c \in \mathbf{c} | c = 0$} \Comment{Check for a rank decrease}
                \State $\rho \gets \rho + \sum_{c \in \mathbf{c}}{\mathcal{I}(c = 0)}$  \Comment{Count spare ranks}
                \State $(\mathbf{A}, \mathbf{B}, \mathbf{c}) \gets f_\textit{prune}(\mathbf{A}, \mathbf{B}, \mathbf{c})$ \Comment{Apply pruning}
            \Else \Comment{Else if not pruned}
                \State $f_\textit{insert}(\Delta\vartheta_u, (\mathbf{A}, \mathbf{B}, \mathbf{c}))$  \Comment{Save adapter for reallocation}
            \EndIf
        \EndFor
        \While {$\rho > 0$} \Comment{While there are spare ranks}
            \For{$(\mathbf{A}, \mathbf{B}, \mathbf{c}) \in \Delta\vartheta_u$} \Comment{Iterate over unpruned adapters}
                \If{$\rho > 0$} \Comment{if there are spare ranks}
                    \State $(\mathbf{A}, \mathbf{B}, \mathbf{c}) \gets f_\textit{reallocate}(\mathbf{A}, \mathbf{B}, \mathbf{c})$ \Comment{Re-allocate a rank}
                    \State $\mathbf{c} \gets \nicefrac{\mathbf{c}}{\sum_{c \in \mathbf{c}}c} $ \Comment{Normalise $\mathbf{c}$ vector}
                    \State $\rho \gets \rho - 1$ \Comment{Update spare ranks}
                \EndIf
            \EndFor
        \EndWhile
    \EndFor
\EndFor\\
\Return $\Delta\vartheta$

\end{algorithmic}

\end{algorithm}

\begin{figure}[!ht] 
\begin{center}
        \subfloat[When a null component in the $\mathbf{c}$ vector of an adapter is detected, the corresponding elements of the adapter are removed using the $f_\textit{prune}(\cdot)$ function, generating a spare rank that will be reallocate. \label{fig:l1rapruning}]{\includegraphics[width=\columnwidth]{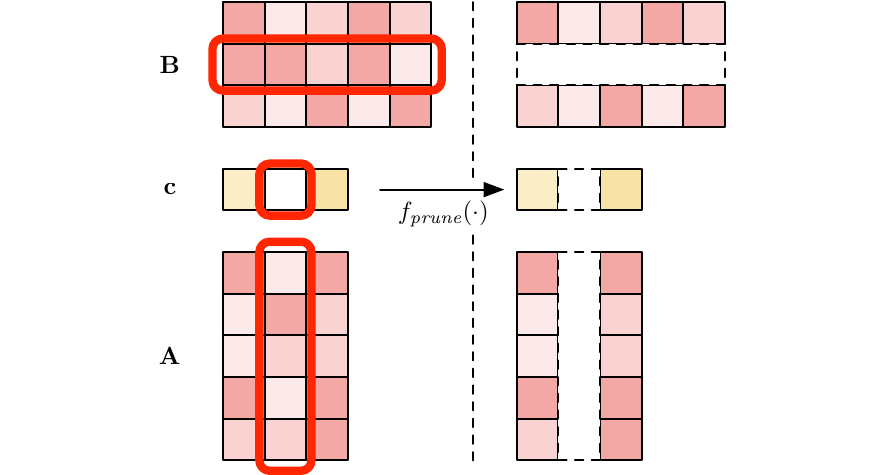}} \\ \vspace{1em}
        \subfloat[When a spare rank is available and a needs to be reallocated, the elements on the target adapter are extended by the $f_\textit{reallocate}(\cdot)$ function (values are initialised as in a regular initialisation). \label{fig:l1rareallocation}]{\includegraphics[width=\columnwidth]{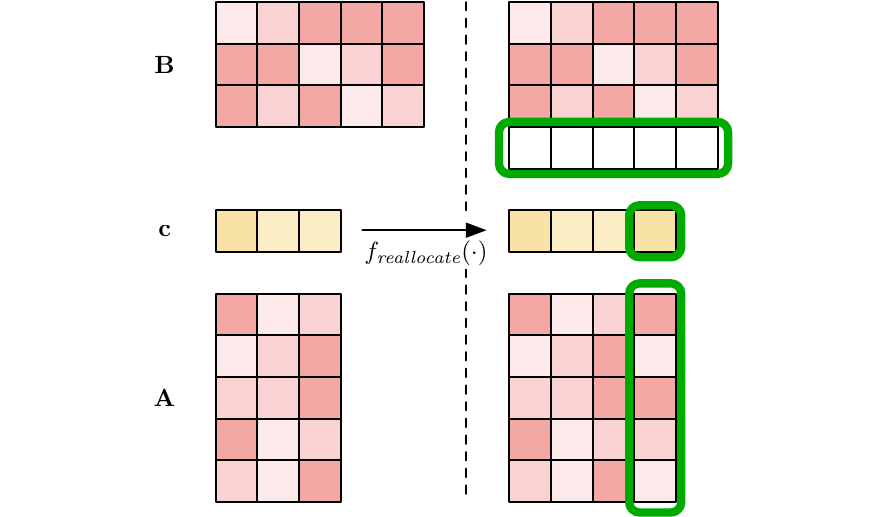}}
    \caption{\textsc{L1RA} pruning and reallocation (lighter colours are for lower absolute values, white is $0$).}
    \label{fig:l1rapruningandreallocation}
\end{center}
\vspace{-1.5em}
\end{figure}

We report the training process of a model using \textsc{L1RA} adapters in the pseudocode detailed in \Cref{alg:trainalg} and we detail the rank pruning and re-allocation process in \Cref{fig:l1rapruningandreallocation}. 
The overall training is similar to that of a model using \textsc{LoRA} adapters. 
The loss function is changed to include the $L_1$ regularisation term (controlled by the $\lambda$ hyperparameter) on the elements of the $\mathbf{c}$ vector. 
Similarly to \textsc{SoRA}, by enforcing sparsity on the $\mathbf{c}$ vector through this regularisation, we achieve rank pruning. 
In fact, whenever an element of $\mathbf{c}$ is shrunk to $0$ the corresponding column in $\mathbf{A}$ and row in $\mathbf{B}$ --the other matrices of the adapter-- can be dropped (this is the role of the $f_\textit{prune}(\cdot)$ function).

All the spare ranks generated by this pruning process can be re-allocated to the other, unpruned, adapters.
Whenever spare ranks are available, the algorithm cycles over the unpruned adapters \emph{sorted by decreasing order of the minimum value in the $\mathbf{c}$ vector}, so that 

\begin{equation}
\label{eq:reallocation_order}
(\mathbf{A}_i, \mathbf{B}_i, \mathbf{c}_i) > (\mathbf{A}_j, \mathbf{B}_j, \mathbf{c}_j) \iff \min{ ~\mathbf{c}_i} > \min{~\mathbf{c}_j}
\end{equation}

\noindent and re-assigns a rank to each adapter until spare ranks are no longer available. In other words, each available additional rank is always redistributed to the particular adapter which is most in need of the rank increase, because its current rank-budget is in full use, with the various components of the $\mathbf{c}$ vector furthest from zero.

The ordering of unpruned adapters is performed in \Cref{alg:trainalg} by the $f_\textit{insert}(\cdot)$ function when saving them in $\Delta\vartheta_u$. 
After the rank re-allocation step, the training procedure reprises.
This sorting step is inspired by SVD: assuming that in high-dimensional space the matrices $\mathbf{A}$ and $\mathbf{B}$ can be treated as orthogonal and the $\mathbf{c}$ vector mimics the diagonal of the singular value matrix.

Compared to other dynamic rank adapters like \textsc{AdaLoRA}, \textsc{SoRA} and \textsc{DyLoRA}, \textsc{L1RA} offers significant advantages. 
If we consider a model using vanilla \textsc{LoRA} adapters with a given rank $r$, since all these other techniques do not account for spare ranks re-allocation, they would require starting from a higher rank initialisation to have the adapters requiring a rank $r' > r$ reach that value, implicitly requiring more memory than the original \textsc{LoRA} would have used.
In contrast, \textsc{L1RA} basically maintains almost the same memory usage by reallocating ranks within the fixed budget (as we detail better in \Cref{sec:discussion}, it cannot always be the same due to some weight matrices having $d_\textit{in}$ or $d_\textit{out}$ different from others). 
Additionally, \textsc{AdaLoRA} increases the requirements on time and memory by imposing SVD behavior to the elements of the adapter through additional terms in the loss function. 
\textsc{L1RA}'s approach avoids these additional constraints, ensuring computational efficiency while achieving performance-optimal rank configuration and maintaining memory limits. 
This makes \textsc{L1RA} a better choice for resource-constrained environments, offering a balanced solution for dynamic rank adaptation.

\section{\textsc{Memory-GELATO}}
\label{sec:appendixb}

\begin{figure*}[!ht]
    \begin{center}
        \includegraphics[width=\textwidth]{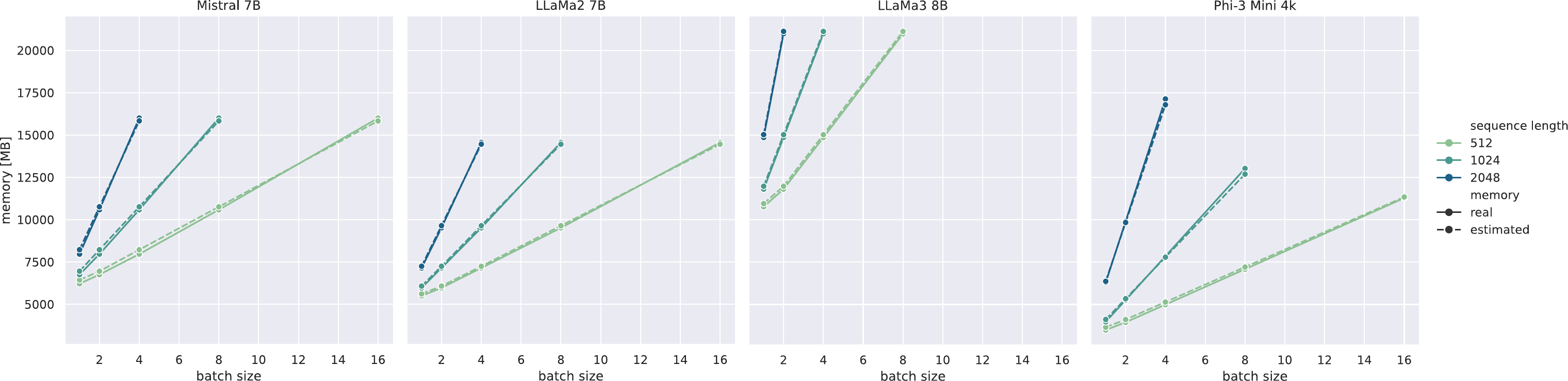}
        \caption{Comparison of peak memory usage estimates from \textsc{Memory-GELATO} against actual peak memory usage during training with \textsc{LoRA} adapters.}
        \label{fig:memgelato}
    \end{center}
\end{figure*}

The \textsc{Memory-GELATO} tool is crucial to reach full memory exploitation.
In fact, it provides an accurate estimate of the memory required to train a model 
We identified the following contribution to memory estimates:
\begin{itemize}
    \item \emph{Model parameters}, which include the weights of all layers and the adapters and is influenced by the numeric precision and quantisation;
    \item \emph{Steady state memory}, that is all memory reserved to keep track of the intermediate states generated by passing data through the model, the gradients and the optimiser state;
    \item \emph{Activation}, that is the additional memory used to memorise the activations for gradient checkpointing (reducing the memory footprint of gradients);
    \item \emph{Loss}, which includes the output logits and the memory used to compute the negative log-likelihood;
    \item \emph{Other contributions}, which includes all the additional elements increasing memory, like operations at the end of the forward pass and the beginning of the backward pass.
\end{itemize}

\begin{table}[ht]
\caption{\textsc{Memory-GELATO} performance in predicting peak memory usage (MAE: Mean Absolute Error, $\rho$:  Spearman correlation coefficient, $r$ Pearson correlation coefficient).}
\begin{center}
\resizebox{\columnwidth}{!}{
\begin{tabular}{cS[table-format=3.2]S[table-format=1.4]S[table-format=1.4]}

\toprule

\textbf{Model} & {\textbf{MAE [MB]}} & $\rho$ & $r$ \\

\midrule

\textsc{Mistral 7B v0.3} & 203.05 & 1.0000 & 0.9998 \\
\textsc{LLaMa2 7B} & 109.80 & 1.0000 & 0.9999 \\
\textsc{LLaMa 3.1 8B} & 159.01 & 1.0000 & 0.9999 \\
\textsc{Phi-3 Mini 4k} & 146.03 & 1.0000 & 0.9998 \\

\bottomrule

\end{tabular}  
}
\end{center}
\label{tab:memgelato}
\end{table}%

To assess the goodness of \textsc{Memory-GELATO} estimates, we compared the predicted and real values of memory peak usage for different models, different maximum sequence lengths and different batch sizes.
In \Cref{fig:memgelato}, we can see the difference between the estimates and the real values; while, in \Cref{tab:memgelato}, we report quantitative metrics on estimates goodness.
Overall, the error in estimated peak memory usage differs from the real one of a few hundreds MBs, including the overestimate we introduced for safety.

\section{Evaluation}
\label{sec:evaluaton}

To evaluate \textsc{L1RA} against other adapter approaches, we applied it to fine-tune a LLM in a realistic use case: assistant fine-tuning.
Moreover, to demonstrate empirically the practical advantages of \textsc{L1RA} against other approaches we used \textsc{Memory-GELATO} to configure the experiment to maximise memory utilisation.
We detail the experimental settings in \Cref{sec:appendixc}.

We experimented fine-tuning to different LLMs (namely \textsc{Mistral 7B v0.3}~\cite{DBLP:journals/corr/abs-2310-06825} and \textsc{LLaMa 3.1 8B}~\cite{DBLP:journals/corr/abs-2407-21783}, both quantised at 4 bits precision) to make sure that \textsc{L1RA} is agnostic of the LLM.
We selected the \textsc{OpenOrca} data set~\cite{DBLP:journals/corr/abs-2306-02707}\footnote{Data set card: \url{https://huggingface.co/datasets/Open-Orca/OpenOrca}} for this assistant fine-tuning. 

In this experiment, we compared the test performance and resource consumption of \textsc{L1RA} against \textsc{LoRA}, \textsc{AdaLoRA}. 
We compared against two versions of \textsc{AdaLoRA}: one targeting the same average rank as \textsc{LoRA} and \textsc{L1RA} starting from an higher rank ($1.5$ times that of \textsc{LoRA} as suggested in the \textsc{AdaLoRA} documentation), and another version starting from the same rank of \textsc{LoRA} and \textsc{L1RA} and targeting a smaller rank (so that the initial one was $1.5$ times that of \textsc{LoRA}, again, as suggested in the \textsc{AdaLoRA} documentation).
This fine-tuning task was chosen to demonstrate the practical application of  \textsc{L1RA} in efficient fine-tuning of LLMs, particularly in scenarios where fine-tuning on consumer-level GPUs is challenging (e.g., when we reach the limit of usable memory).  

Throughout the experiment, we kept track of ranks evolution to analyse the final distribution at the end of training.
It this way we can get a better understanding of which components within the Transformer architecture need a more precise adaptation (identified as those with a higher adapter rank) and shed lights on the internal mechanisms of the Transformer architecture.

\label{sec:experiments}

\section{Results}
\label{sec:results}

\begin{table*}[!htp]
\caption{Results and resources consuption of chatbot assistant fine-tuning on the \textsc{OpenOrca} data set (\emph{Rank}: is the initial adapters rank --for \textsc{AdaLoRA} we have the initial rank and target average--, \emph{PPL}: perplexity --lower is better, bold values are the best result for each model--, \emph{Time}: total time for training, validation and testing; \emph{Memory}: peak VRAM usage during training; \emph{No. of adapter parameters}: trainable adapter parameters at the end of training).}
\begin{center}
\resizebox{.975\textwidth}{!}{
\begin{threeparttable}[b]
\begin{tabular}{cc S[table-format=2.0] S[table-format=1.2] S[table-format=5.2]S[table-format=2.2] S[table-format=2.2]S[table-format=2.2]}

\toprule

\multirow{2}{*}{\textbf{Model}} & \multirow{2}{*}{\textbf{Approach}} & {\multirow{2}{*}{\textbf{Rank}}} & {\multirow{2}{*}{\textbf{PPL} $\downarrow$}} & {\multirow{2}{*}{\textbf{Training time [s]} $\downarrow$}} & {\multirow{2}{*}{\textbf{Memory [GB]}\tnote{1} $\downarrow$}} & \multicolumn{2}{c}{\textbf{No. of adapter parameters [M]} $\downarrow$} \\ \cmidrule(l){7-8}
& & & & & & {\textbf{Start of training}} & {\textbf{End of training}} \\

\midrule

\multirow{4}{*}{\textsc{LLaMa 3 8B}} & \textsc{LoRA} & 16 & 3.32 & 30994.89 & 13.84 & 41.94 & 41.94 \\
& \textsc{AdaLoRA} & {$24 \rightarrow 16$} & 3.63{\tnote{2}} & 32980.28 & 14.00 & 62.92 & 62.92 \\
& \textsc{AdaLoRA} & {$16 \rightarrow 12$} & 3.57{\tnote{2}} & 32964.14 & 13.76 & 41.95 & 41.95 \\
& \textsc{L1RA} & 16 & $\mathbf{3.25}$ & 31246.40 & 14.23 & 41.95 & 45.16 \\

\midrule

\multirow{4}{*}{\textsc{Mistral 7B v0.3}} & \textsc{LoRA} & 16 & 2.93 & 37891.88 & 13.58 & 41.94 & 41.94 \\
& \textsc{AdaLoRA} & {$24 \rightarrow 16$} & 3.16{\tnote{2}} & 40234.87 & 13.82 & 62.92 & 62.92 \\
& \textsc{AdaLoRA} & {$16 \rightarrow 12$} & 3.16{\tnote{2}} & 40215.02 & 13.59 & 41.95 & 41.95 \\
& \textsc{L1RA} & 16 & $\mathbf{2.91}$ & 37968.91 & 13.94 & 41.95 & 50.06 \\

\bottomrule

\end{tabular}  

\begin{tablenotes}
\item [1] Values measured using \textsc{PyTorch} utility for measuring GPU device memory usage: \url{https://pytorch.org/docs/stable/generated/torch.cuda.max_memory_allocated.html}.
\item[2] Values are slightly altered because PPL was computed from the loss of the model which included also the regularisation term, separate computations showed that \textsc{AdaLoRA} PPL was higher than that of \textsc{LoRA} and \textsc{L1RA}.
\end{tablenotes}

\end{threeparttable}
}
\end{center}
\label{tab:chatbotresults}
\end{table*}%

\begin{table*}[!htp]
\caption{Relative results and resources consumption from \Cref{tab:chatbotresults} normalised to the \textsc{LoRA} fine-tuning.}
\begin{center}
\resizebox{\textwidth}{!}{
\begin{threeparttable}[b]
\begin{tabular}{cc S[table-format=2.0] S[table-format=1.2] S[table-format=1.2]S[table-format=1.2] S[table-format=2.2]}

\toprule

\textbf{Model} & \textbf{Approach} & {\textbf{Rank}} & {\textbf{$\bm{\Delta}$ PPL [$\%$]} $\downarrow$} & {\textbf{$\bm{\Delta}$ Training time [$\%$]} $\downarrow$} & {\textbf{$\bm{\Delta}$ Memory [$\%$]} $\downarrow$} & {\makecell{\textbf{$\bm{\Delta}$ No. of adapter parameters [$\%$]}\tnote{1} $\downarrow$}} \\

\midrule

\multirow{3}{*}{\textsc{LLaMa 3 8B}} & \textsc{AdaLoRA} & {$24 \rightarrow 16$} & +9.34 & +6.41 & +1.16 & +50.02 \\
& \textsc{AdaLoRA} & {$16 \rightarrow 12$} & +7.53 & +6.35 & -0.58 & +0.02 \\
& \textsc{L1RA} & 16 & -2.11 & +0.81 & +2.82 & +7.68 \\

\midrule

\multirow{3}{*}{\makecell[c]{\textsc{Mistral}\\\textsc{7B v0.3}}} & \textsc{AdaLoRA} & {$24 \rightarrow 16$} & +7.85 & +6.18 & +1.77 & +50.02 \\
& \textsc{AdaLoRA} & {$16 \rightarrow 12$} & +7.85 & +6.13 & +0.07 & +0.02 \\
& \textsc{L1RA} & 16 & -0.68 & +0.20 & +2.65 & +19.36 \\

\bottomrule

\end{tabular}  

\begin{tablenotes}
\item [1] Values computed on end-of-training parameters.
\end{tablenotes}

\end{threeparttable}
}
\end{center}
\label{tab:relativeresources}
\end{table*}%

\begin{figure}[!ht] 
\begin{center}
    \subfloat[\textsc{LLaMa 3.1 8B}. \label{fig:rankevollama}]{\includegraphics[width=.9\columnwidth]{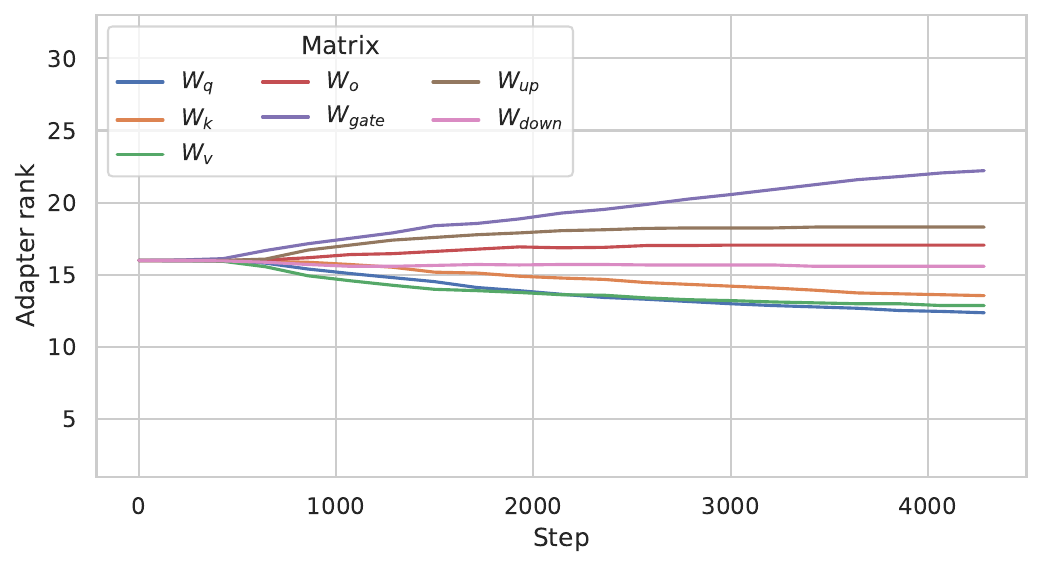}} \\
    \subfloat[\textsc{Mistral 7B v0.3}. \label{fig:rankevomistral}]{\includegraphics[width=.9\columnwidth]{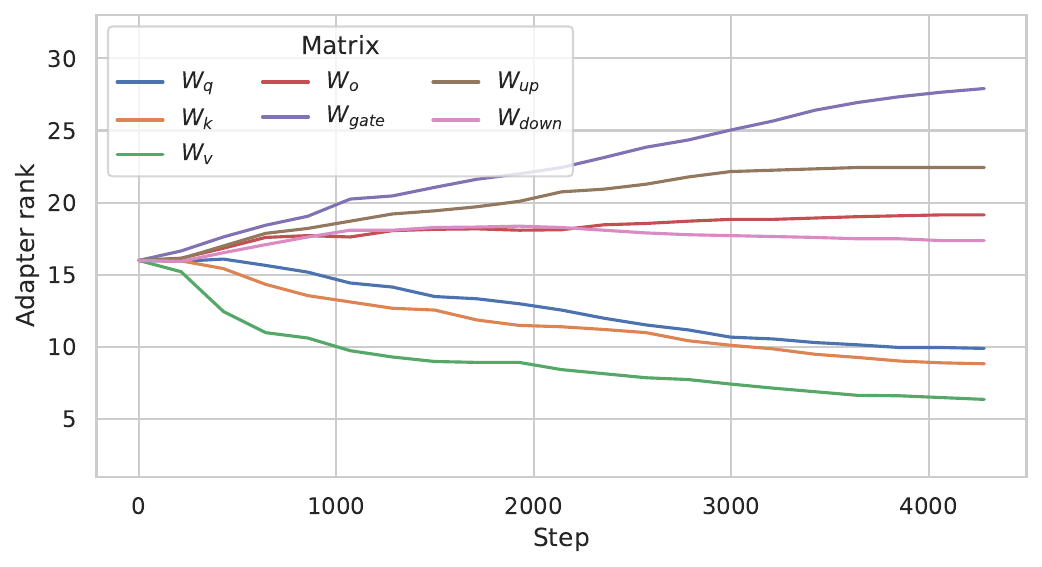}}
    \caption{Matrix-wise evolution of layer-wise average \textsc{L1RA} adapters rank during training.}
\label{fig:rankevo}
\end{center}
\end{figure}

\begin{figure}[!ht] 
\begin{center}
    \includegraphics[width=.7\columnwidth]{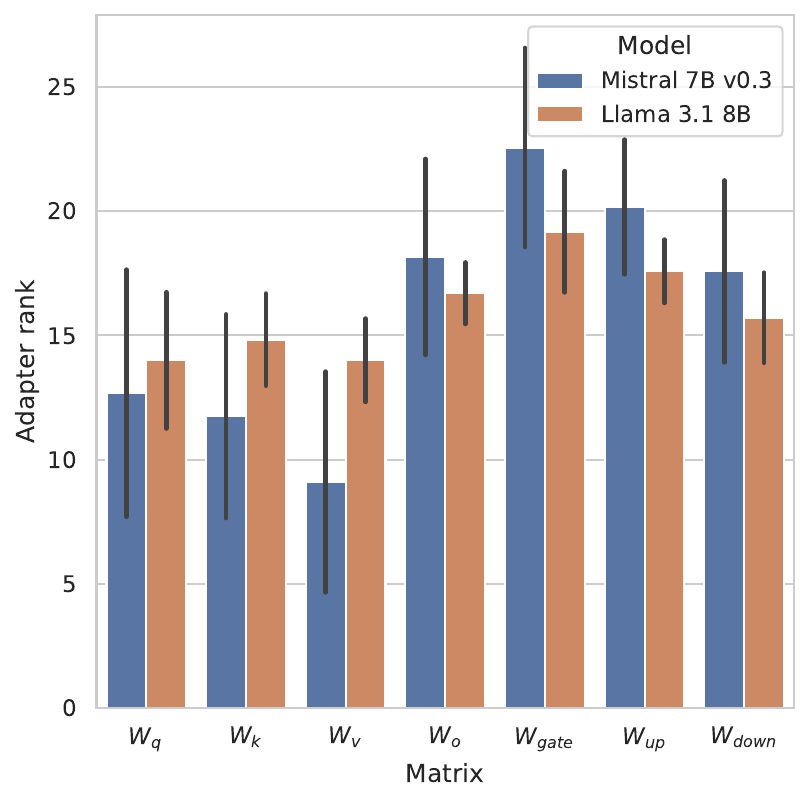}
    \caption{Matrix-wise distribution of layer-wise average \textsc{L1RA} adapters rank at the end of training (error bars show standard deviation).}
\label{fig:rankdist}
\end{center}
\end{figure}

\begin{figure}[!ht] 
\begin{center}
    \includegraphics[width=.9\columnwidth]{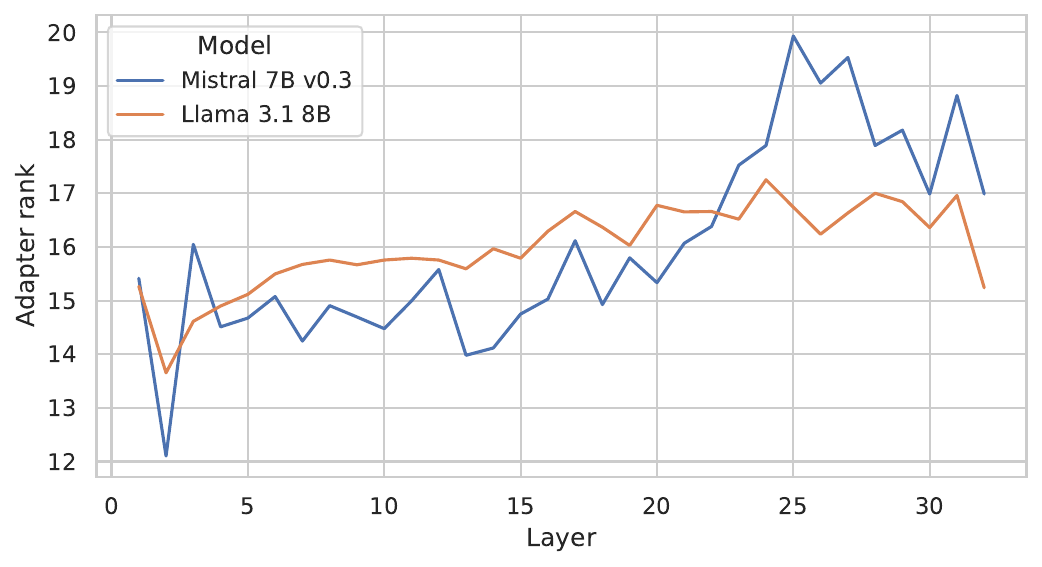}
    \caption{Layer-wise evolution of matrix-wise average \textsc{L1RA} adapters rank at the end of training.}
\label{fig:reassignedranks}
\end{center}
\vspace{-1.5em}
\end{figure}

\begin{figure*}[!ht] 
\begin{center}
    \subfloat[\textsc{LLaMa 3.1 8B} start of training. \label{fig:detailedrankevollamastart}]{\includegraphics[width=\columnwidth]{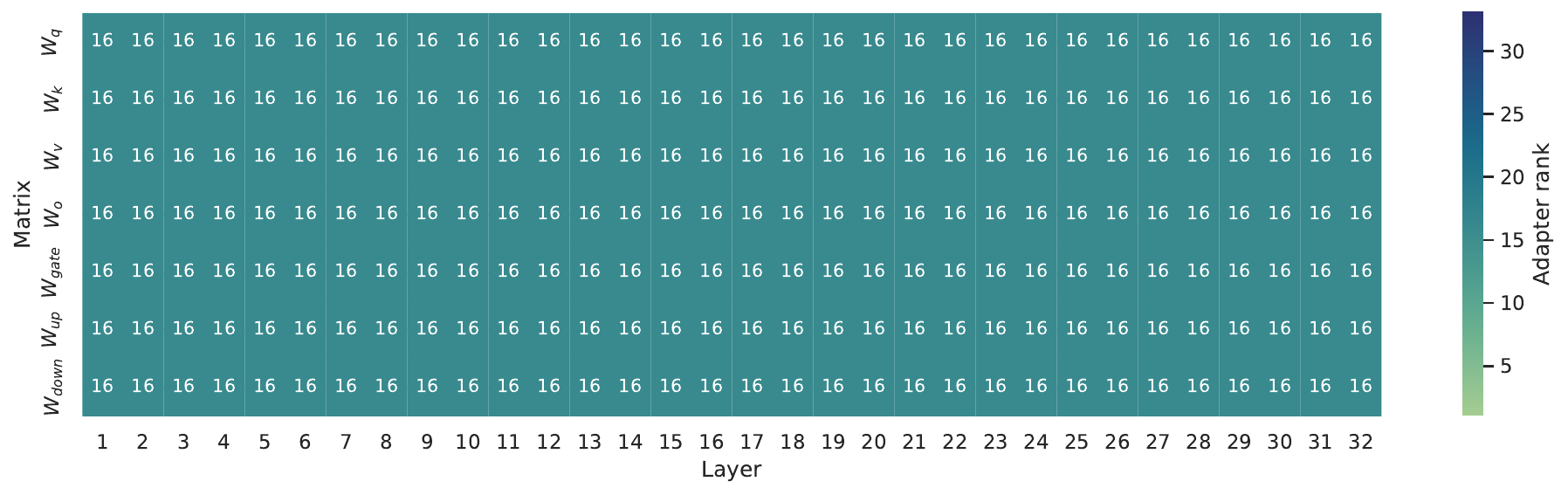}} \hfill
    \subfloat[\textsc{Mistral 7B v0.3} start of training. \label{fig:detailedrankevomistrastartl}]{\includegraphics[width=\columnwidth]{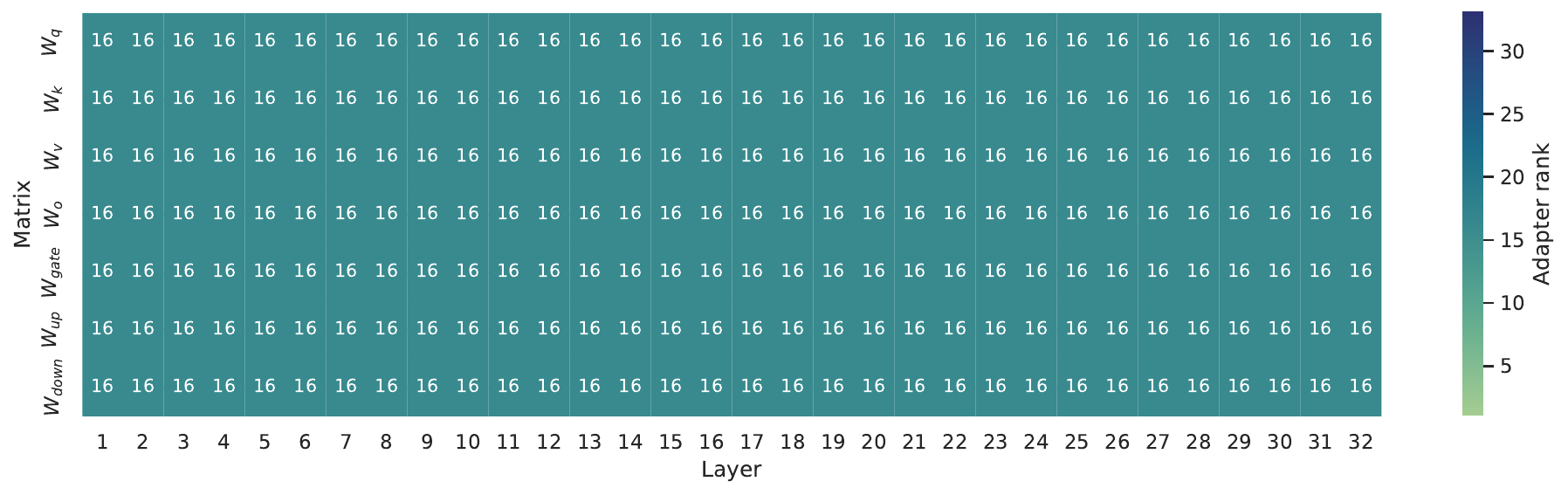}} \\
    \subfloat[\textsc{LLaMa 3.1 8B} halfway through training. \label{fig:detailedrankevollamamid}]{\includegraphics[width=\columnwidth]{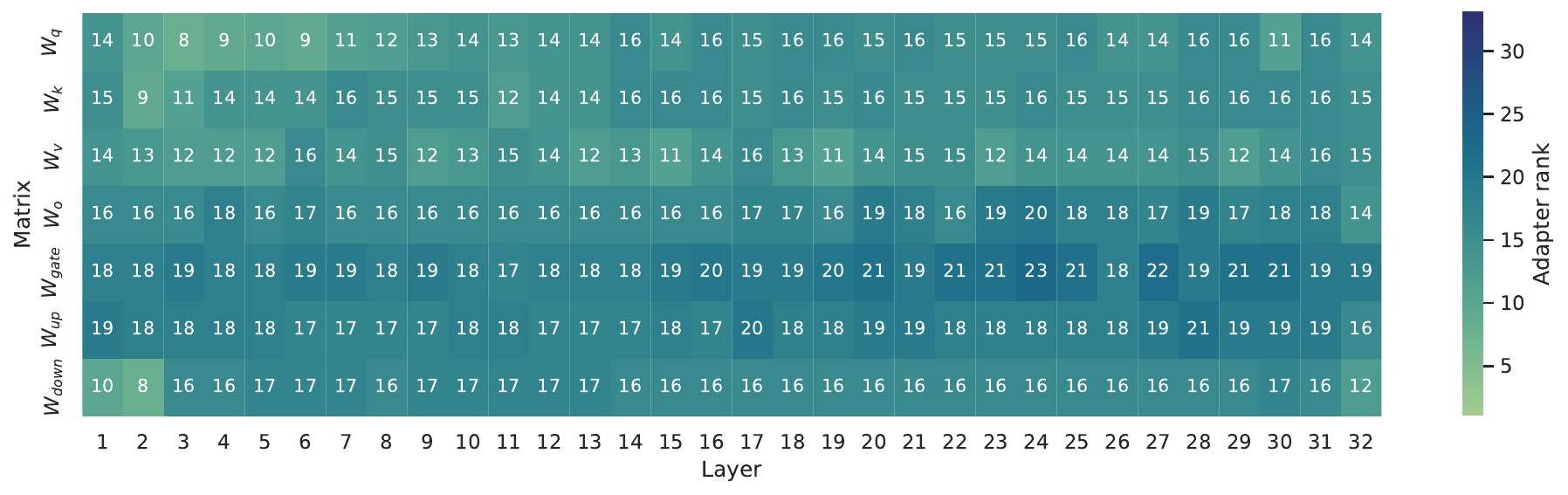}} \hfill
    \subfloat[\textsc{Mistral 7B v0.3} halfway through training. \label{fig:detailedrankevomistralmid}]{\includegraphics[width=\columnwidth]{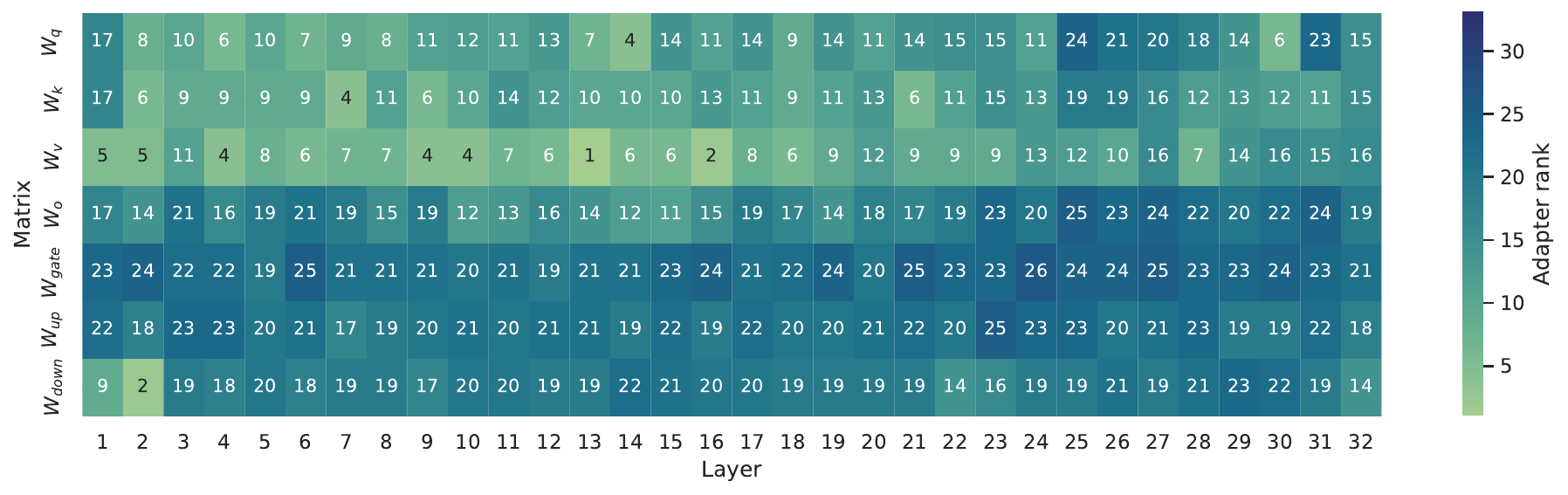}} \\
    \subfloat[\textsc{LLaMa 3.1 8B} end of training. \label{fig:detailedrankevollamaend}]{\includegraphics[width=\columnwidth]{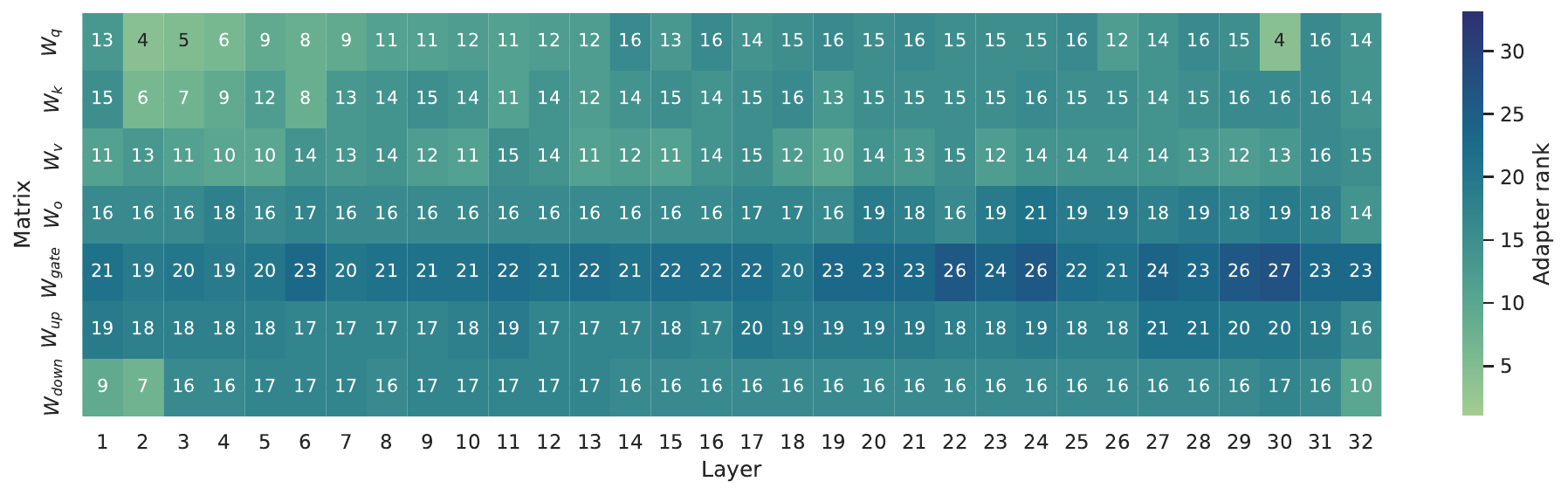}} \hfill
    \subfloat[\textsc{Mistral 7B v0.3} end of training. \label{fig:detailedrankevomistralend}]{\includegraphics[width=\columnwidth]{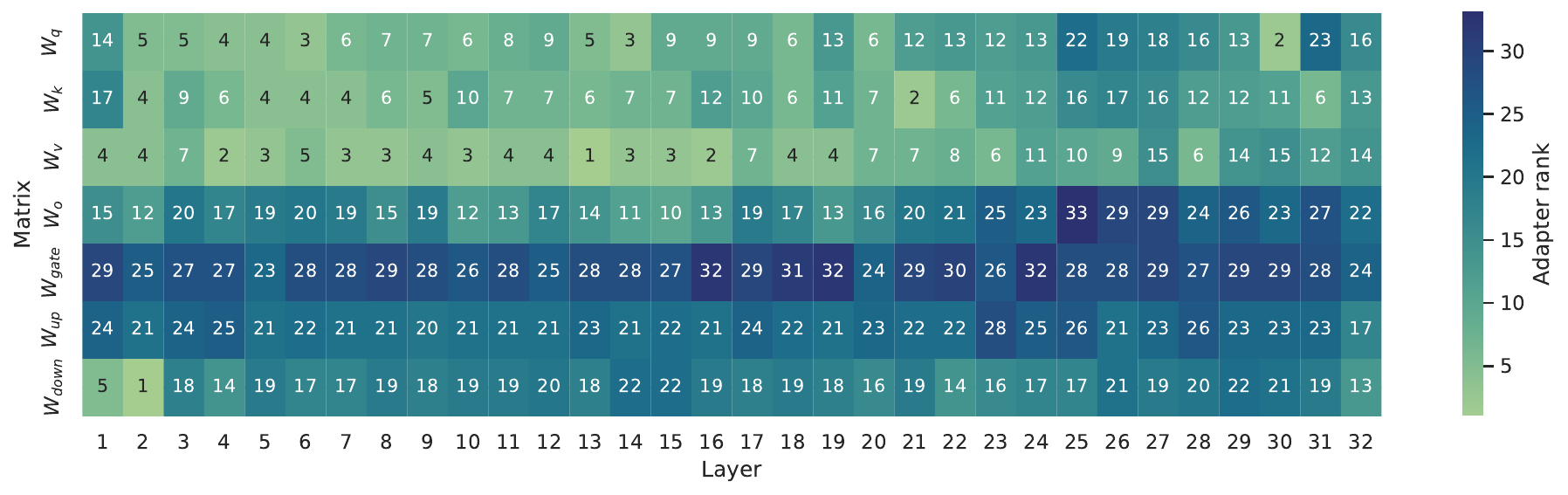}}
    \caption{Layer-wise and Matrix-wise evolution of \textsc{L1RA} adapters rank during training.}
\label{fig:detailedrankevo}
\end{center}
\vspace{-1.5em}
\end{figure*}

We report the main results of the experiments in \Cref{tab:chatbotresults}, while the relative values, to ease the comparison, are in \Cref{tab:relativeresources}.
\textsc{L1RA} achieves the lowest absolute perplexity (PPL) score, improving over both \textsc{LoRA} and \textsc{AdaLoRA}.
Moreover, \textsc{L1RA} achieves also the closest training time to that of \textsc{LoRA}, with less than $1\%$ difference from \textsc{LoRA}.
Memory consumption, on the other hand, seems to be similar among the three approaches (most differences from \textsc{LoRA} are below $2\%$) with \textsc{AdaLoRA} performing better than \textsc{L1RA} (and even \textsc{LoRA} in one configuration).
The number of adapter (trainable) parameters shows how \textsc{AdaLoRA} not applying an actual pruning the matrices and requiring an higher starting rank to target the same average ranks of \textsc{LoRA} and \textsc{L1RA} increases significantly the number of parameters ($50\%$) without an improvement on the PPL. 
On the other side, \textsc{L1RA} exchanging freely parameters between matrices of different sizes causes an increase in the number of parameters as training continues; however, the increase is smaller than that of \textsc{AdaLoRA} and achieves better PPL than both \textsc{LoRA} and \textsc{AdaLoRA}.
We discuss better about memory consumption and number of adapter parameters in \Cref{sec:discussion}.

In \Cref{fig:rankevo} we can the average evolution of the ranks organised per matrix of the Transformer architecture.
As we can see, \textsc{LLaMa} and \textsc{Mistral} have the same trends: matrices coming from the \emph{Feed-Forward Neural Network} (FFNN) layer of the Transformer architecture (up-projection $\mathbf{W}_\textit{up}$, gate $\mathbf{W}_\textit{gate}$, and down-projection $\mathbf{W}_\textit{down}$) are more ``rank hungry'' than those of the \emph{multi-head self-attention} layers (key $\mathbf{W}_k$, value $\mathbf{W}_v$, query $\mathbf{W}_q$,  and output $\mathbf{W}_o$).
At the end of training, the difference is clear across all layers, as shown by the averaged rank counts in \Cref{fig:rankdist}.
From \Cref{fig:reassignedranks} we can see another common trend between the two models: ranks are higher in the layers closer to the output of the neural network.

Finally, to report on the individual ranks of each matrix in the Transformer stack, we can see in \Cref{fig:detailedrankevo} ranks distributions at the beginning, halfway through and at the end of training.
The darker area emerging at the bottom corresponds to the the matrices of the FFNN.
We can see how the ``rank mass'' is higher in these layers especially toward the top of the Transformer network (bottom-right side on the plot) and how it is lower for the multi-head self-attention layer matrices at the bottom of the Transformer network.
Moreover, we can wee how with \textsc{LLaMa} this trend is emerging slower: the matrix showing rank distribution at the end is closer to that of \textsc{Mistral} halfway through training.
Given the higher PPL, we can assume that \textsc{LLaMa} could have been trained for more iterations.

\section{Discussion}
\label{sec:discussion}

Values of memory consumption does not comply with our expectations, especially if compared taking into account the rank distributions and the number of trainable parameters.
Considering the average ranks and the total number of parameters, we expected to see \textsc{AdaLoRA} starting from the higher rank having the highest memory consumption and \textsc{AdaLoRA} starting from the same rank as \textsc{LoRA} still consume more memory due to the additional operations to compute the regularisation term, while the memory is even lower in the case of \textsc{LLaMa}.
We suspect this is due to some internal optimisation or offloading of the trainer in the \textsc{HuggingFace}'s \textsc{Transformers} library we used~\cite{DBLP:conf/emnlp/WolfDSCDMCRLFDS20}.
Despite we were not able to locate the source of this difference, we conducted a small experiments on the same data using the same handmade training loop with all adapters and we measured an overall higher memory consumption that was more in line with the number of parameters and the compared techniques.
In the next iteration of \textsc{L1RA} we plan to drop the trainer to have more reliable estimates.

To comment on the difference in number of parameters between \textsc{L1RA} and \textsc{LoRA}, we can see that despite \textsc{L1RA} not exceeding the rank budget, the amount of parameters (and used memory) is slightly higher than \textsc{LoRA}.
This is a result of allocating the spare ranks to other adapters working on matrices of different sizes.
In particular, as we saw from \Cref{fig:rankdist}, many ranks are allocated to the feed-forward layers, which have a higher ($4 \times$) inner projection dimensionality.
Despite this situation, \textsc{L1RA} still achieves a lower resources utilisation when compared to \textsc{AdaLoRA}.

Finally, to comment on the trends observed in \Cref{fig:rankevo,fig:rankdist,fig:reassignedranks,fig:detailedrankevo}, we can say that trends hint how the FFNN layers at the top of the Transformer stack are contributing more to the task being solved.
The high-level features processed in that part of the Transformer network need more precise refinement thus the higher rank.
Similarly, we believe that exploiting different information from other tokens in the context is not as important as extracting more refined patterns with the non-linear transformations of the FFNN to have the LLM behave as a chatbot assistant, thus the higher ranks in FFNN layers.
This observation agrees with intuition that the higher layers of the network should contribute the most to adapting the network to a specific domain, and that the output and FFNN layers are crucial for storing domain-specific information (as noted by \cite{DBLP:conf/emnlp/GevaSBL21,DBLP:conf/nips/BidermanPSSAPR23}) that likely needs to be updated by the adapters.

\section{Conclusion}
\label{sec:conclusion}

In this paper, we introduced \textsc{L1RA}, a novel technique for efficient LLM fine-tuning. 
By effectively exploiting the dynamic rank assignment given by $L_1$ regularisation and re-assigning the spare ranks within the available budget, \textsc{L1RA} represents a significant advancement in efficient fine-tuning, offering a promising solution for resource-constrained environments. 
We completed \textsc{L1RA} with \textsc{Memory-GELATO} our tool for GPU memory estimation we can exploit to determine the memory --and thus rank-- budget.
At this moment we foresee two possible, complementary, directions in the further development of \textsc{L1RA}: we are interested in studying the rank distribution across different models and at a different scales or number of parameter and data-set sizes and we are interested in better understanding the convergence of the proposed method.

\section*{Limitations}

In this paper, we mainly focused on the development of \textsc{L1RA} for efficient fine-tuning and its evaluation on realistic use cases, rather than exhaustive experiments.
The first limitation is in the choice of the LLM: as for now, we evaluated the results using only \textsc{Mistral 7B v0.3} and \textsc{LLaMa 3 8B}. A proper evaluation would require exploring other openly accessible models of the same and different sizes that would fit on a consumer-level GPU.
The second limitation is the choice of the evaluation data set: we considered only the task of instruction fine-tuning since it is a common use case and since it covers many tasks an LLM is required to solve, however a more extensive evaluation exploring different tasks would improve the understanding of \textsc{L1RA}'s capabilities.

\section*{Ethics Statement}

The authors do not foresee any considerable risks associated with the work presented in this paper. 
In principle, the \textsc{L1RA} algorithm is intended to make fine-tuning of LLMs more efficient and the \textsc{Memory-GELATO} tool is intended for estimating memory consumption such fine-tunings.
The authors made the source code publicly available to ensure the reproducibility of the experiments.
Refer to \Cref{sec:appendixa} for further details.

\section*{Acknowledgements}

This work was partially supported by the FAIR (Future Artificial Intelligence Research) project, funded by the NextGenerationEU program within the PNRR-PE-AI scheme (M4C2, Investment 1.3, Line on Artificial Intelligence), by funding from the pilot program Core Informatics at KIT (KiKIT) of the Helmholtz Association (HGF), and supported by the German Research Foundation (DFG) - SFB 1608 - 501798263 and KASTEL Security Research Labs, Karlsruhe.

\bibliography{anthology,custom}
\bibliographystyle{acl_natbib}

\clearpage

\appendix

\section{Source Code Availability}
\label{sec:appendixa}

We share the source code associated with this paper for full reproducibility and transparency. 
All the source material to replicate the experiments is available on GitHub:
\begin{itemize}
    \item \textsc{L1RA}: \url{https://github.com/raul-singh/L1RA/tree/dev-exp};
    \item \textsc{Memory-GELATO}: \url{https://github.com/raul-singh/memory-gelato}.
\end{itemize}

\section{Evaluation setup}
\label{sec:appendixc}

In this section, we provide the hyperparamers we used for in experimental evaluations to ensure full reproducibility.
We report the hyperparameters we used with the \textsc{OpenOrca} data set in \Cref{tab:guanacoconfigs}.
In the table, we use the following notation:
\begin{itemize}
    \item $r$ is the initial rank of \textsc{L1RA}, \textsc{LoRA} and \textsc{AdaLoRA} adapters;
    \item $\alpha$ is the scaling of \textsc{L1RA}, \textsc{LoRA} and \textsc{AdaLoRA};
    \item $p_\textit{dropout}$ is the dropout probability of \textsc{L1RA}, \textsc{LoRA} and \textsc{AdaLoRA};
    \item $\eta$ is the learning rate;
    \item $\lambda$ is the regularisation coefficient or \textsc{L1RA} or \textsc{AdaLoRA};
\end{itemize}


\begin{table}[!ht]
\caption{\textsc{OpenOrca} hyperparameters.}
\begin{center}
\resizebox{\columnwidth}{!}{
\begin{tabular}{ccc}

\toprule

\textbf{Model} & \textbf{Hyperparameter} & {\textbf{Value}} \\

\midrule

\multirow{20}{*}{\makecell[c]{\textsc{LLaMa} \\ \textsc{3.1 8B}}} & Max. sequence length & $1024$ tokens \\
& $r$ & $16$ \\
& $\alpha$ & $16$ \\
& $p_\textit{dropout}$ & $10^{-1}$ \\
& Compute d-type & \texttt{bfloat16} \\
& Attn. implementation & Flash attn. 2~\cite{DBLP:conf/iclr/Dao24} \\ \cmidrule(l){2-3}

& Optimiser & AdamE (paged, 32 bit) \\
& $\eta$ & $10^{-4}$ \\
& $\eta$ schedule & cosine \\
& $\eta$ warm-up ratio & $5\%$ \\
& Max grad. norm & $1$ \\
& Epochs & $1$ \\
& Batch size & $4$ \\
& Accum. steps & $4$ \\ \cmidrule(l){2-3}

& $\lambda_\textsc{L1RA}$ & $10^{-3}$ \\
& $\eta_\mathbf{c}$ (\textsc{L1RA}) & $10^{-2}$ \\
& Rank update period (\textsc{L1RA}) & $5\%$ training steps \\
& $\lambda_\textsc{AdaLoRA}$ & $10^{-3}$ \\
& $t_\textit{init}$ (\textsc{AdaLoRA}) & $5\%$ training steps \\
& $\Delta t$ (\textsc{AdaLoRA}) & $5\%$ training steps \\

\midrule

\multirow{20}{*}{\makecell[c]{\textsc{Mistral} \\ \textsc{7B v0.3}}} & Max. sequence length & $1024$ tokens \\
& $r$ & $16$ \\
& $\alpha$ & $16$ \\
& $p_\textit{dropout}$ & $10^{-1}$ \\
& Compute d-type & \texttt{bfloat16} \\
& Attn. implementation & Flash attn. 2~\cite{DBLP:conf/iclr/Dao24} \\ \cmidrule(l){2-3}

& Optimiser & AdamE (paged, 32 bit) \\
& $\eta$ & $10^{-4}$ \\
& $\eta$ schedule & cosine \\
& $\eta$ warm-up ratio & $5\%$ \\
& Max grad. norm & $1$ \\
& Epochs & $1$ \\
& Batch size & $4$ \\
& Accum. steps & $4$ \\ \cmidrule(l){2-3}

& $\lambda_\textsc{L1RA}$ & $10^{-3}$ \\
& $\eta_\mathbf{c}$ (\textsc{L1RA}) & $10^{-2}$ \\
& Rank update period (\textsc{L1RA}) & $5\%$ training steps \\
& $\lambda_\textsc{AdaLoRA}$ & $10^{-3}$ \\
& $t_\textit{init}$ (\textsc{AdaLoRA}) & $5\%$ training steps \\
& $\Delta t$ (\textsc{AdaLoRA}) & $5\%$ training steps \\

\bottomrule

\end{tabular}  
}
\end{center}
\label{tab:guanacoconfigs}
\end{table}%

One important detail of our experiments is the choice of the optimiser, we implemented a variant of \emph{AdamW}~\cite{DBLP:conf/iclr/LoshchilovH19} (which is the most common optimiser for LLMs), to support decoupled regularisation for both $L_1$ and $L_2$ regularisations. 
We refer to this variant as \emph{AdamE}, where the ``E'' refers to \emph{ElasticNet}: the combined $L_1$ and $L_2$ regulariser\footnote{AdamE implementation \url{https://github.com/vincenzo-scotti/bitsandbytes/tree/dev-adame}}.
The addition is the decoupled $L_1$ regularisation that avoids the update of the lasso constraint being scaled by the adaptive learning rate and momentum hyperparameters.
This scaling affects negatively the shrinking of the parameters, showing it down.

Since we apply learning rate warm-up and cosine scheduling to shrink $\eta$ to zero, we find useful keep a separate constant learning rate for the parameters in the $\mathbf{c}$ vectors.
To avoid introducing unnecessary hyperaparameters we use the same $\eta$ of the rest of the parameters, but whithout warm-up and scheduling.

We conducted all experiments on the same machine with the following hardware configuration:
\begin{itemize}
    \item CPU: Intel Core i9-13900K;
    \item RAM: 64 GB;
    \item GPU: NVIDIA GeForce RTX 4090.
\end{itemize}

We used as much shared parameters across the three approaches we compare (\textsc{L1RA}, \textsc{LoRA} and \textsc{AdaLoRA}) as possible to have a fair comparison.

\end{document}